\pdfoutput=1
\documentclass[letterpaper]{article} 
\usepackage{aaai23}  
\usepackage{times}  
\usepackage{helvet}  
\usepackage{courier}  
\usepackage[hyphens]{url}  
\usepackage{graphicx} 
\usepackage{natbib}  
\usepackage{caption} 
\usepackage{subfigure}
\usepackage{amsmath}
\usepackage{amssymb}
\usepackage{multirow}

\usepackage{algorithm}
\usepackage{algorithmic}

\usepackage{newfloat}
\usepackage{listings}
\DeclareCaptionStyle{ruled}{labelfont=normalfont,labelsep=colon,strut=off} 
\floatstyle{ruled}
\newfloat{listing}{tb}{lst}{}
\floatname{listing}{Listing}
\title{Exploring Stroke-Level Modifications for Scene Text Editing}
\author{
    Yadong Qu,\textsuperscript{\rm 1}
    Qingfeng Tan,\textsuperscript{\rm 2}\thanks{Corresponding author.}
    Hongtao Xie,\textsuperscript{\rm 1}
    Jianjun Xu,\textsuperscript{\rm 1}
    Yuxin Wang,\textsuperscript{\rm 1}
    Yongdong Zhang\textsuperscript{\rm 1}
}
\affiliations{
    \textsuperscript{\rm 1} University of Science and Technology of China, \textsuperscript{\rm 2} GuangZhou University\\

    \{qqqyd, xujj1998, wangyx58\}@mail.ustc.edu.cn, tqf528@gzhu.edu.cn, \{htxie, zhyd73\}@ustc.edu.cn
}

\usepackage{bibentry}

\begin{document}

\maketitle

\begin{abstract}
Scene text editing (STE) aims to replace text with the desired one while preserving background and styles of the original text.
However, due to the complicated background textures and various text styles, existing methods fall short in generating clear and legible edited text images.
In this study, we attribute the poor editing performance to two problems:
1) Implicit decoupling structure. Previous methods of editing the whole image have to learn different translation rules of background and text regions simultaneously.
2) Domain gap. Due to the lack of edited real scene text images, the network can only be well trained on synthetic pairs and performs poorly on real-world images.
To handle the above problems, we propose a novel network by MOdifying Scene Text image at strokE Level (MOSTEL).
Firstly, we generate stroke guidance maps to explicitly indicate regions to be edited.
Different from the implicit one by directly modifying all the pixels at image level, such explicit instructions filter out the distractions from background and guide the network to focus on editing rules of text regions.
Secondly, we propose a Semi-supervised Hybrid Learning to train the network with both labeled synthetic images and unpaired real scene text images.
Thus, the STE model is adapted to real-world datasets distributions.
Moreover, two new datasets (Tamper-Syn2k and Tamper-Scene) are proposed to fill the blank of public evaluation datasets.
Extensive experiments demonstrate that our MOSTEL outperforms previous methods both qualitatively and quantitatively.
Datasets and code will be available at https://github.com/qqqyd/MOSTEL.
\end{abstract}

\section{Introduction}
As an emerging task in recent years, scene text editing (STE) has received increasing attention.
It is designed to replace text in a scene image with new text while maintaining the original background textures and text styles (\emph{e.g.} font, color, size, spatial transformation, etc.).
STE can convert any word in an image to a desired one within a second and retain visual consistency before and after tampering, eliminating the need to spend hours manually editing images.
The edited images can be used as augmented data to train the scene text detector~\cite{qin2021mask,qu2022adnet} and recognizer~\cite{sheng2019nrtr,du2022SVTR}.
Compared with existing synthesis methods~\cite{jaderberg2014synthetic}, which simply place font-specific text on random background images, STE can better simulate various text styles and provide more reliable training images. 
It can also be used in sensitive data protection, smart city~\cite{qiu2019automatic} and augmented reality translation~\cite{fragoso2011translatar}.

\begin{figure}[tbp]
	\centering
	\subfigbottomskip=0pt
	\subfigcapskip=0pt
	\subfigure[]{
		\includegraphics[width=\linewidth]{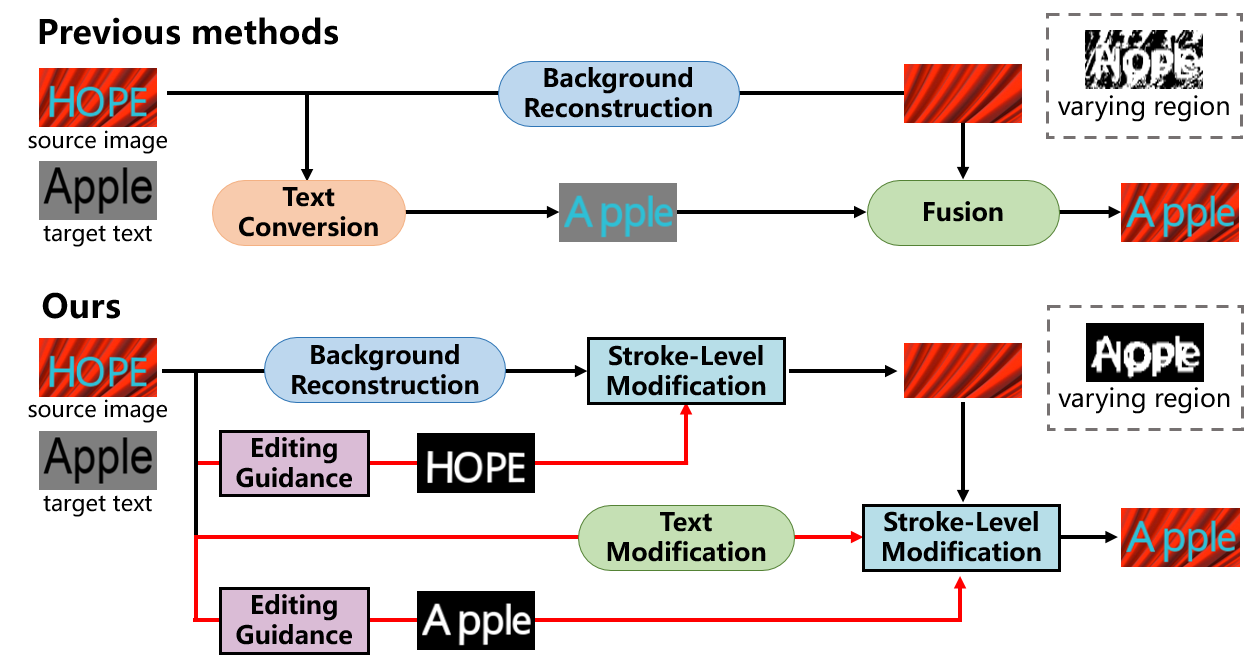}
		\label{fig:fig1a}
	}
	\\
	\subfigure[]{
		\includegraphics[width=\linewidth]{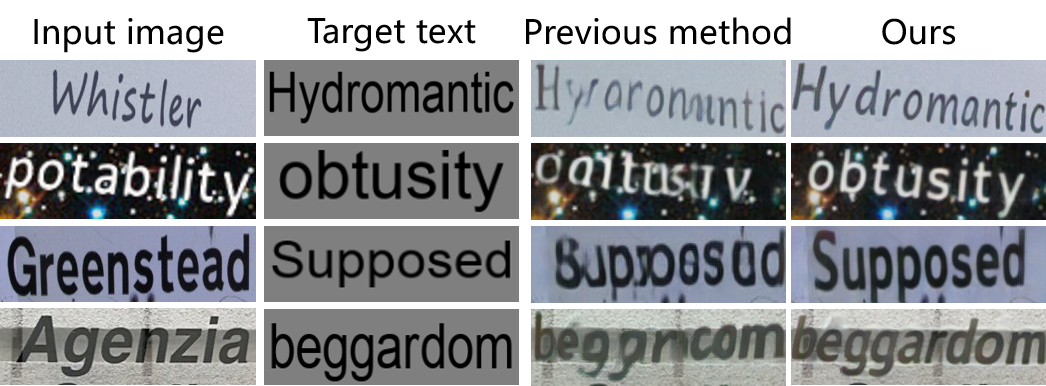}
		\label{fig:fig1b}
	}
	\caption{The comparison between MOSTEL and previous methods. (a) is the pipeline. The dashed boxes indicate all the changed pixels throughout the editing process. (b) is the qualitative examples of previous method and ours.}
	\label{fig:fig1}
\end{figure}

STE task has two main difficulties: 1) It is challenging to simulate various text styles while keeping the background textures intact.
2) The dearth of real-world training pairs causes the domain gap between synthetic training data and real scene text images.
As shown in Fig. \ref{fig:fig1a}, previous methods~\cite{srnet,swaptext} divide the STE task into three simpler subtasks: background reconstruction, text conversion and fusion.
However, they address STE as a vanilla image-to-image translation task and directly use labeled synthetic images to train these modules.
This implicit guidance by modifying all the pixels at image level aggravates the learning difficulty of editing rules.
While learning how to edit texts, networks still strive to distinguish and preserve the regions that need to be retained.
This way of implementing two tasks in a single module distracts the network from concentrating on the editing rules in text regions, resulting in illegible typeface and poor imitation of styles (Fig. \ref{fig:fig1b}).
In addition, the divide-and-conquer methods require supervision on each sub-network.
The absence of corresponding labels on real-world images allows the network only to be trained with synthetic images, causing domain bias in real scene text images during test stage.

In this paper, we take a further step towards accurate scene text editing and propose a novel framework by MOdifying Scene Text image at strokE Level (MOSTEL).
First, as shown in Fig. \ref{fig:fig1a}, we propose the editing guidance to predict the guidance maps to indicate the regions to be edited.
They guide stroke-level modifications by explicitly filtering out invariant background regions.
Compared with previous implicit guidance, our explicit one has two advantages.
On the one hand, the editing guidance explicitly decomposes the modification of text regions and the maintenance of background regions, enabling the network to focus on the editing rules of text regions.
By eliminating the distractions from invariant background, it considerably decreases the learning difficulty of editing rules and ensures the consistency of generated text styles with the original.
On the other hand, as shown in the dash boxes in Fig. \ref{fig:fig1a}, previous implicit methods still cause subtle changes to the background regions.
Since the invariant background is directly inherited from the source image, MOSTEL can maintain the integrity of the original image background to the greatest extent.

Second, a Semi-supervised Hybrid Learning is proposed to bridge the domain gap between training data and real-world test data.
In the training stage, we introduce the unpaired scene text images by converting the text to itself.
Specifically, we adopt the erase-and-write paradigm and propose Background Reconstruction Module and Text Modification Module to remove the unnecessary intermediate results.
The real-world training images are first erased to generate text-free background, and then the same text with imitated style is rewritten onto the reconstructed background.
In such a training scheme, to avoid the model degenerating into an identity mapping network, several countermeasures are adopted to separate the background and text, which are described in detail in Methodology.
Therefore, we successfully adapt STE model to real-world scene text datasets distributions.
Moreover, a scene text recognizer is employed in the training stage to ensure the generated text images clear and legible.

In addition, two new STE datasets named Tamper-Syn2k and Tamper-Scene are proposed to evaluate the performance of scene text editors.
As far as we know, they are the first publicly available STE evaluation datasets, which will significantly facilitate a fair comparison of STE methods and promote the development of STE task.
Extensive experiments on these datasets also demonstrate the superiority of our method both quantitatively and qualitatively.

Our contributions are summarized as follows:
\begin{itemize}
	\item{We propose a novel framework to perform stroke-level modifications, which explicitly guides the network to focus on the learning of editing rules in text regions and maximizes the integrity of background regions.}
	\item{We design a Semi-supervised Hybrid Learning that enables the model to be trained using both labeled synthetic images and unpaired real-world images.}
	\item{Two new STE datasets (Tamper-Syn2k and Tamper-Scene) are released to ensure a fair comparison of STE methods and promote the development of STE task.}
	\item{MOSTEL achieves promising performance in both quality and quantity. Our simple and powerful model will provide many insights for future STE works.}
\end{itemize}

\section{Related Work}
\subsection{Text Image Synthesis}
Text image synthesis is a major trick for training robust DNN models.
Several attempts have been made to generate synthetic text images in order to improve the accuracy of text detection and recognition. 
For example, ~\cite{jaderberg2014synthetic} use a word generator to insert texts into semantically sensible regions of background images. 
\cite{zhan2018vis} take semantic coherence, visual attention and adaptive text appearance into account to generate realistic synthetic text images. 
Most recently, text image synthesis methods have advanced significantly as a result of the development of Variational Auto Encoder (VAE) ~\cite{kingma2013vae} and Generative Adversarial Networks (GAN)~\cite{goodfellow2014gan}. 
\cite{yang2019controllable} achieve real-time control of the crucial stylistic degree of the glyph via an adjustable parameter. 
\cite{sun2017scvae} use a VAE structure to implement a stylized Chinese character generator.
Meanwhile, ~\cite{azadi2018mcgan} propose an end-to-end solution to synthesize ornamented glyphs from images of several reference glyphs in the same style.

\subsection{Style Transfer}
\begin{figure*}[t]
	\centering
	\includegraphics[width=0.92\textwidth]{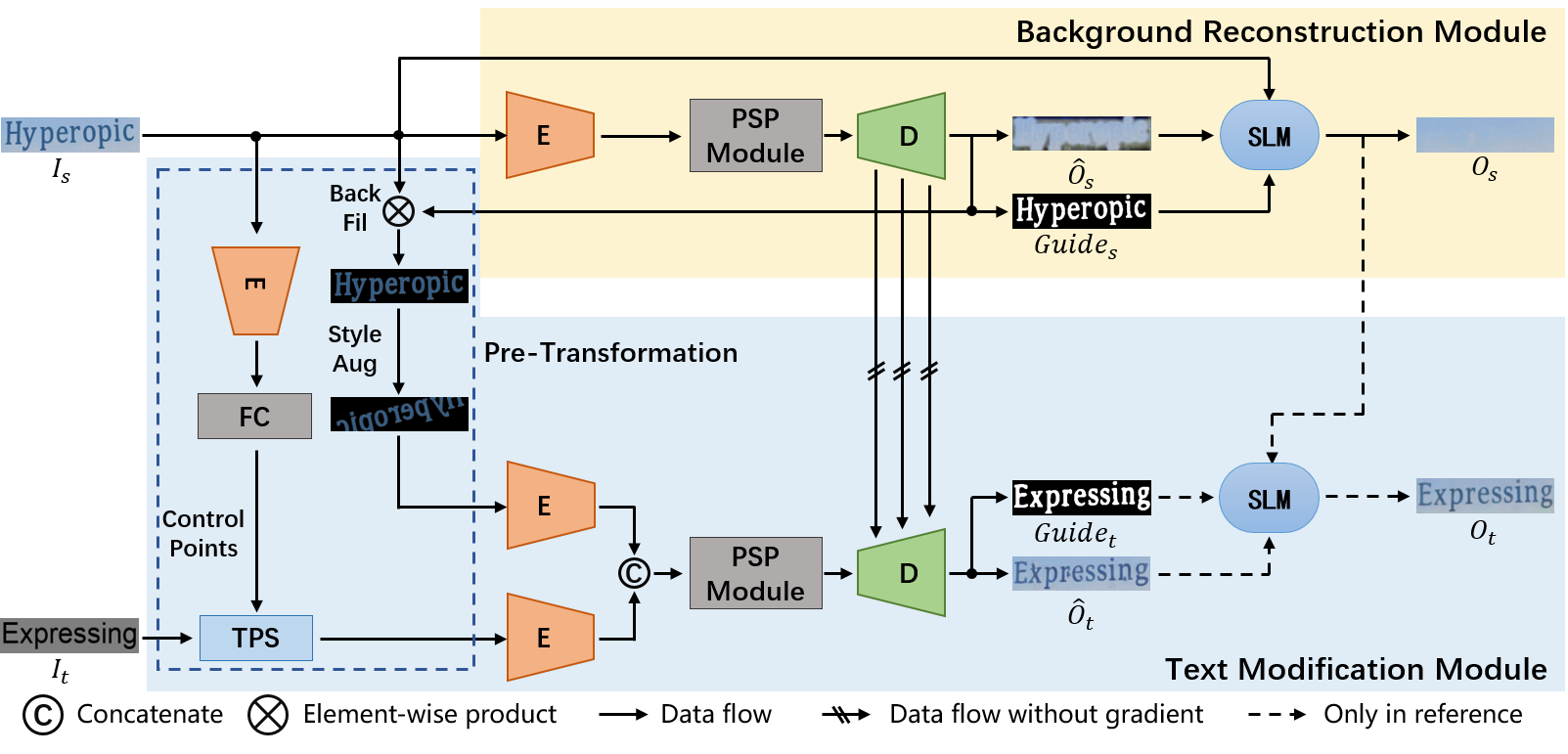}
	\caption{The overall pipeline of MOSTEL. The network consists of Background Reconstruction Module and Text Modification Module. Back Fil, Style Aug and SLM mean Background Filtering, Style Augmentation and Stroke-Level Modification.}
	\label{fig:pipeline}
\end{figure*}

Image style transfer is another challenging related task that aims to transfer visual style from a reference image to another target image. 
Most existing methods apply the encoder-decoder architecture that embeds the input into a subspace and then decodes it to generate desired images. 
\cite{isola2017pixel2pixel} implement a learnable mapping from the input image to the output image using a training set of aligned image pairs. 
\cite{zhu2017cyclegan} introduce cycle consistency loss to generalize the mapping relationship to unpaired cross-domain data. 
Similar ideas have been applied to various tasks, such as generating photos from sketches and face synthesis from attribute and semantic information. 
\cite{yang2017awesome_typography} are the first to apply style transfer methods to text images. 
They analyze and model the distance-based essential characteristics of text effects and leverage them to guide the synthesis process.
Meanwhile, ~\cite{yang2019tetgan} accomplish both the objective of style transfer and style removal by using stylization and de-stylization sub-networks.

\subsection{Scene Text Editing}
Due to the wide range of applications, GAN-based scene text editing methods attract increasing research interest.
STEFANN~\cite{stefann} designs a Font Adaptive Neural Network to edit a single character.
However, this character-level modification fails to replace a word with length changes, limiting performance in practical applications. 
SRNet~\cite{srnet} first proposes the word-level editing method by dividing the network into three sub-networks: background inpainting, text conversion and fusion.
The divide-and-conquer strategy allows each module to handle only a relatively simple task.
Extended on SRNet, SwapText~\cite{swaptext} introduces the TPS module that isolates the spatial transformation from text styles, reducing the learning difficulty of text conversion module.
\cite{zhao2021deep} apply scene text editing to document forgery and propose the forge-and-recapture operation to mitigate the visual artifacts.
STRIVE~\cite{strive} employs SRNet for video text editing.
They only modify a selected frame as reference and add targeted photometric transformations to the reference frame to maintain the consistency of all other modified frames.
These approaches, however, are mainly extended from SRNet, which may fail to replace a text with complex styles and can only be trained on synthetic datasets.
To deal with these issues, we propose a stroke-level modification method to generate more legible text images.
Our method also supports the semi-supervised training scheme and can be trained on both labeled synthetic datasets and unpaired scene text images.

\section{Methodology}
The proposed MOSTEL aims to generate legible scene text images and maintain the background integrity as much as possible.
To this end, we follow the erase-and-write paradigm and design a novel stroke-level modification network.
As the overall pipeline in Fig. \ref{fig:pipeline} shows, MOSTEL is composed of Background Reconstruction Module (BRM) and Text Modification Module (TMM).
BRM takes the source image $I_s$ as input and outputs the stroke guidance map ${Guide}_{s}$ and text-free background $O_s$.
In TMM, input pairs, source image $I_s$ and standard targeted text image $I_t$, are first processed by Pre-Transformation.
To be specific, a TPS module is used to adjust the orientation of $I_t$ according to the geometrical attributes of $I_s$.
$I_s$ is processed by Background Filtering and Style Augmentation, which are supposed to filter out redundant background textures and preserve independent and robust text style information.
Then the transformed pairs are fused with reconstructed background features to generate the edited text image.
Stroke-Level Modification is implemented in both BRM and TMM to ensure more reliable outputs. 
Details will be introduced in the following sections.

\subsection{Background Reconstruction Module}
Background Reconstruction Module (BRM) aims to generate text-free background images with proper textures.
Inspired by SRNet~\cite{srnet}, BRM adopts an encoder-decoder structure.
The encoder consists of three down-sampling layers and four residual blocks, and the decoder consists of three up-sampling layers.
To obtain a more robust feature representation, a PSP Module~\cite{zhao2017pyramid} is applied to enhance multi-scale features of the encoder.
However, addressing scene text editing as a vanilla image-to-image translation problem is suboptimal and has two shortcomings.
First, while the background is expected to remain the same, directly modifying all the pixels at image level still causes subtle, imperceptible changes to the background regions.
Second, the invariant background distracts the model from focusing on the varying text regions and hinders the learning of editing rules.

Therefore, inspired by PERT~\cite{wang2021pert}, we propose a structure equipped with Stroke-Level Modification (SLM) to perform explicit guiding editing by minimally modifying the varying regions.
The prediction of editing guidance maps is considered a segmentation task of stroke regions.
When generating the partially reconstructed image $\hat{O}_s$, the network also predicts editing guidance map ${Guide}_{s}$.
Then $\hat{O}_s$ and ${Guide}_{s}$ are fed into SLM to obtain the text-free background image $O_s$.
This process can be formulated as:
\begin{equation}\label{eq:slm}
	O_s = {Guide}_{s} \times \hat{O}_s + (1 - {Guide}_{s}) \times I_s.
\end{equation}

This explicit guidance brought by SLM can generate more reliable edited images.
On the one hand, background regions are directly inherited from source image $I_s$, maximizing the consistency with invariant regions of the source image.
On the other hand, the model can get rid of distinguishing the invariant and varying regions and only focus on the editing rules of text regions, thus simplifying the task and facilitating the learning.
SLM is further quantitatively proved to be beneficial in ablation experiments.

The text-free background image $O_s$ is optimized with GAN loss and $L2$ loss.
\begin{equation}\label{eq:loss_o_s}
	\begin{aligned}
		L_{b\_s} = \mathbb{E}[logD_b&(T_b, I_s) + log(1 - D_b(O_b, I_s))] + \\
		&\lambda_{b1} \Vert T_b - O_b \Vert_2,
	\end{aligned}
\end{equation}
where $T_b$ and $O_b$ are the ground truth and predicted background images.
$\lambda_{b1}$ is set to 10 to balance numeric values.
$D_b$ indicates the background discriminator, which follows the structure in SRNet~\cite{srnet}.
The supervision on the editing guidance ${Guide}_{s}$ adopts dice loss, which can be formulated as:
\begin{equation}\label{eq:loss_b_mask_s}
	\begin{aligned}
		L_{b\_guide} = 1 - \frac{2\left|T_{guide\_s}\cap {Guide}_{s}\right|}{\left|T_{guide\_s}\right| + \left|{Guide}_{s}\right|},
	\end{aligned}
\end{equation}
where $T_{guide\_s}$ and ${Guide}_{s}$ are the ground truth and predicted guidance map.
$\left| X\right|$ means the sum of pixels in X.

\subsection{Text Modification Module}
Text Modification Module (TMM) is composed of two parts: Pre-Transformation and Modification Module.
They are used to preprocess input images and mix text styles with background features to generate the edited image, respectively.

\subsubsection{Pre-Transformation} applies a specialized spatial transformation to the input pairs, source image $I_s$ and standard targeted text image $I_t$.
Inspired by Swaptext~\cite{swaptext}, a feature extractor and two FC layers are used to obtain control points of $I_s$, which are several anchor points located on contour of the text.
According to the predicted control points, a TPS module is used to geometrically transform $I_t$ into the same orientation as $I_s$.
By decoupling spatial attributes from the other text styles, it reduces the style transfer difficulty.
As for $I_s$, because the jumbled background textures only distract the model from learning text styles, the editing guidance ${Guide}_{s}$ is first introduced to remove background noises in Background Filtering.
To generate more robust style features, Style Augmentation is proposed to enhance text styles by applying random rotation and flipping. 
Since the spatial orientation is decomposed by the TPS module, rotating and flipping operations in Style Augmentation cause no effect on the other styles (such as font, color, size).
When the proposed Semi-supervised Hybrid Learning is used to train the model on real-world scene text images, Pre-Transformation also plays a crucial role in preventing the network from turning into an identity mapping one, which will be described in detail below.

\subsubsection{Modification Module} adopts the same encoder-decoder structure as BRM.
In the decoder, the targeted text features are fused with the corresponding up-sampling features from BRM to generate the edited text image $\hat{O}_t$ and guidance map $Guide_t$.
The gradient of connections is blocked, which is another measure to avoid the network falling into identity mapping when adopting Semi-supervised Hybrid Learning.
It is worth mentioning that we only apply SLM in the inference stage here.
This is because SLM aims to explicitly guide the network to concentrate on text regions and avoid distractions from invariant background.
Pre-Transformation, which may also be regarded as the stroke-level modification, has filtered out background regions in advance, making SLM less important here.
We further provide experiments to verify the effectiveness of such a structure.

In addition, a pre-trained text recognizer is introduced to ensure the generated text image clear and legible.
The recognizer can use any advanced scene text recognition method.
In our implementation, considering the trade-off between performance and speed, we adopt ~\cite{baek2019wrong} as our recognizer, which is made up of TPS transformation, BiLSTM decoder and attention-based prediction.

The loss function on edited scene text image $\hat{O}_t$ uses both GAN loss and $L2$ loss.
\begin{equation}\label{eq:loss_o_t}
	\begin{aligned}
		L_{t\_t} = \mathbb{E}[logD_t(T_t, &I_t) + log(1 - D_t(\hat{O}_t, I_t))] + \\
		&\lambda_{t1} \Vert T_t - \hat{O}_t \Vert_2,
	\end{aligned}
\end{equation}
where $T_t$ and $\hat{O}_t$ are the ground truth and predicted edited images.
$\lambda_{t1}$ is the balance factor and set to 10.
The edited image discriminator $D_t$ has the same structure as $D_b$.
The recognizer uses cross-entropy loss.
\begin{equation}\label{eq:loss_rec}
	\begin{aligned}
		L_{rec} = -\frac{1}{N} \sum_{i=1}^{N}log(p_i \vert g_i),
	\end{aligned}
\end{equation}
where $p_i$ and $g_i$ represent the prediction and ground truth.
$N$ indicates the maximum prediction length and is set to 25.
As with $Guide_s$, the supervision on $Guide_t$ also uses dice loss.
\begin{equation}\label{eq:loss_t_mask_t}
	\begin{aligned}
		L_{t\_guide} = 1 - \frac{2\left|T_{guide\_t}\cap {Guide}_{t}\right|}{\left|T_{guide\_t}\right| + \left|{Guide}_{t}\right|}.
	\end{aligned}
\end{equation}

To generate more realistic images, VGG-loss is adopted, which is divided into a perceptual loss ~\cite{johnson2016perceptual} and a style loss ~\cite{gatys2016image}.
\begin{equation}\label{eq:loss_vgg}
	\begin{aligned}
		L_{vgg} = \lambda_{v1}L_{per} + \lambda_{v2}L_{style},
	\end{aligned}
\end{equation}
\begin{equation}\label{eq:loss_per}
	\begin{aligned}
		L_{per} = \mathbb{E}[\Vert \phi_i(T_t) - \phi_i(\hat{O}_t)   \Vert_1 ],
	\end{aligned}
\end{equation}
\begin{equation}\label{eq:loss_style}
	\begin{aligned}
		L_{style} = \mathbb{E}_j[\Vert G_j^\phi(T_t) - G_j^\phi(\hat{O}_t) \Vert_1 ],
	\end{aligned}
\end{equation}
where $\phi_i$ is the activation map from \emph{relu1\_1} and \emph{relu5\_1} layer of VGG-19 model.
$G$ is the Gram matrix, and balance factors $\lambda_{v1}$ and $\lambda_{v2}$ are set to 1 and 500, respectively.

The whole loss function can be expressed as
\begin{equation}\label{eq:loss_t}
	\begin{aligned}
		L = \arg \mathop{\min}\limits_{G} \mathop{\max}\limits_{D_b, D_f} (& L_{b\_s} + L_{t\_t} + \lambda_1(L_{b\_guide} + \\
		& L_{t\_guide}) + \lambda_2 L_{vgg} + \lambda_3 L_{rec}),
	\end{aligned}
\end{equation}
where the weighting factors $\lambda_1$, $\lambda_2$ and $\lambda_3$ are set to 10, 1 and 0.1.

\subsection{Semi-supervised Hybrid Learning}
\begin{figure}[tbp]
	\centering
	\includegraphics[width=\linewidth]{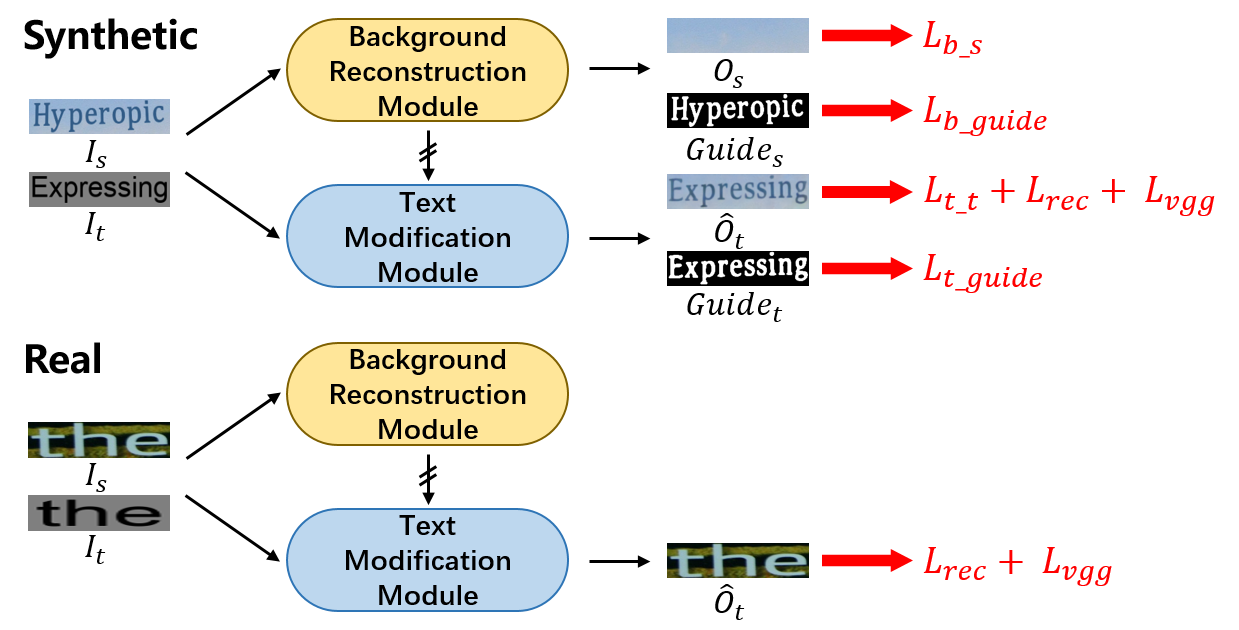}
	\caption{Illustration of Semi-supervised Hybrid Learning. Top and bottom are the training processes on labeled synthetic images and unpaired scene text images, respectively.}
	\label{fig:scheme}
\end{figure}

To adapt the model to real-world environments, we propose the Semi-supervised Hybrid Learning (SHL).
In such a scheme, only transcript annotations are needed, which can be easily accessed in scene text recognition datasets~\cite{icdar2013,icdar2015}.

As shown in Fig. \ref{fig:scheme}, the network structure is the same when training on synthetic and real-world images.
Due to the lack of required labels in real scene text images, we design a paradigm by converting the text to itself.
To be specific, the source image $I_s$ is first processed by BRM to generate text-free background.
Standard targeted text image $I_t$ still imitates styles from $I_s$ and is then fused with erased background features from BRM.
There are some changes in the loss functions.
The supervision on guidance maps ${Guide}_{s}$, ${Guide}_{t}$ and reconstructed background image $O_s$ are discarded.
The loss function for edited image $\hat{O}_t$ is made up by only $L_{rec}$ and $L_{vgg}$.

However, if the network directly uses SHL, because the supervision is the input image itself, the model may degenerate into an identity mapping network, resulting in disastrous performance in the inference stage.
Different from existing methods~\cite{srnet,swaptext}, several operations are employed to prevent this collapse.
As we stated before, Pre-Transformation is firstly used to perform the spatial transformation on the input images.
With the background-free and augmented text styles, it is difficult for the network to find a simple correspondence between input and output images.
Moreover, when fused with background features in TMM, we stop the gradient of the connections from BRM, that is, prevent the encoder and decoder in BRM from restoring the text style information.
Through this specialized designed structure, we separate the background and text style apart in BRM and TMM, respectively.
Extensive experiments further demonstrate the effectiveness of the network structure and the semi-supervised training scheme.

\section{Experiment}
\subsection{Datasets}
The datasets used for training and evaluation are introduced as follows.
To our knowledge, there are no public evaluation datasets for scene text editing.
Therefore, we release two new datasets, Tamper-Syn2k and Tamper-Scene, for a fair comparison between STE methods.
\subsubsection{Synthetic Data.}
We generate 150k labeled images for the supervised training for MOSTEL and 2k paired images to compose Tamper-Syn2k for evaluation.
The paired images are rendered with different texts and the same styles, such as font, size, color, spatial transformation and background image.
In our implementation, a total of 300 fonts and 12,000 background images are used with random rotation, curve and perspective transformation.

\subsubsection{Real Data.}
We use MLT-2017\footnote{https://rrc.cvc.uab.es/?ch=8} to train MOSTEL on real-world scene text images, including 34,625 images.
Only transcript annotations are required to indicate the targeted text and calculate recognizer loss.
For evaluation, the proposed Tamper-Scene is a combination of several scene text datasets, including ICDAR 2013~\cite{icdar2013}, SVT~\cite{svt}, SVTP~\cite{svtp}, IIIT~\cite{iiit5k}, MLT-2019\footnote{https://rrc.cvc.uab.es/?ch=15}, and COCO-Text~\cite{coco-text}.
By filtering the severely distorted and unrecognizable images, we select a total of 7,725 images to compose Tamper-Scene.

\subsection{Implementation Details}
Style Augmentation in Pre-Transformation includes random rotation with an angle from $-15^\circ$ to $15^\circ$ and random flipping with a probability of 0.5 during the training stage.
Input images are resized to $256 \times 64$.
We adopt Adam optimizer with $\beta_1 = 0.9$ and $\beta_2 = 0.999$, and learning rate is set to $5 \times 10^{-5}$.
We totally train 300k iterations with a batch size of 16, consisting of 14 labeled synthetic image pairs and 2 unannotated real scene text images.
MOSTEL is implemented in PyTorch and trained on 1 NVIDIA 2080Ti GPU.

\subsection{Evaluation Metrics}
To comprehensively evaluate the edited images of our method, for paired synthetic images, we adopt the following commonly used metrics: 1). MSE, the $L2$ distances; 2). PSNR, the ratio of peak signal to noise; 3). SSIM, the mean structural similarity; 4) FID~\cite{heusel2017gans}, the distances between features extracted by InceptionV3.
A higher PSNR, SSIM and lower MSE, FID indicate better performance.
For real scene text images, we adopt the recognition accuracy using an official text recognition algorithm~\cite{baek2019wrong} with its corresponding pre-trained model\footnote{https://github.com/clovaai/deep-text-recognition-benchmark}.

\subsection{Ablation Study}
\begin{figure}[tbp]
	\centering
	\includegraphics[width=0.95\linewidth]{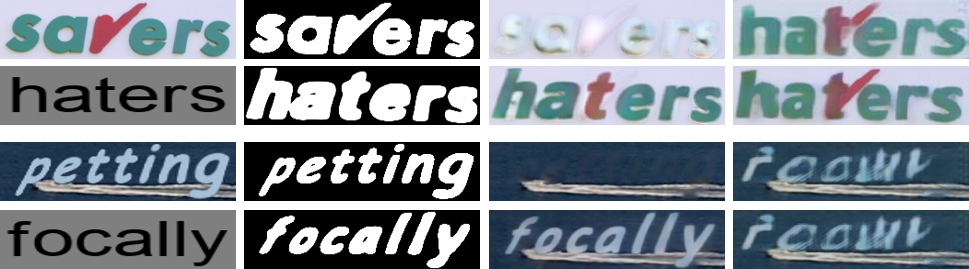}
	\caption{Two visual examples of guidance maps. From left to right, from top to bottom are $I_s$, ${Guide}_{s}$, $O_s$, result of SRNet, $I_t$, ${Guide}_{t}$, $O_t$, result of Swaptext.}
	\label{fig:vis1}
\end{figure}

\begin{figure}[tbp]
	\centering
	\includegraphics[width=0.95\linewidth]{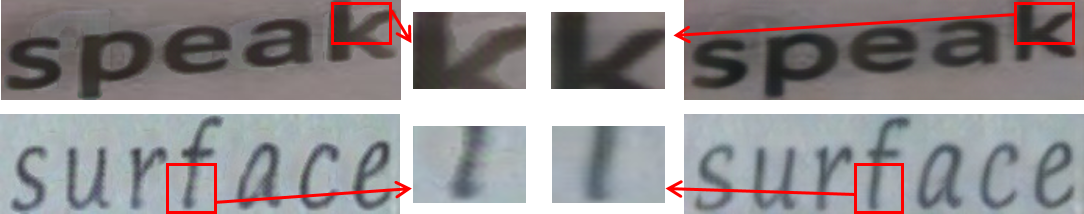}
	\caption{Left and right are the results of adopting SLM in TMM during training and inference stage.}
	\label{fig:slm}
\end{figure}

\begin{figure*}[tbp]
	\centering
	\includegraphics[width=\textwidth]{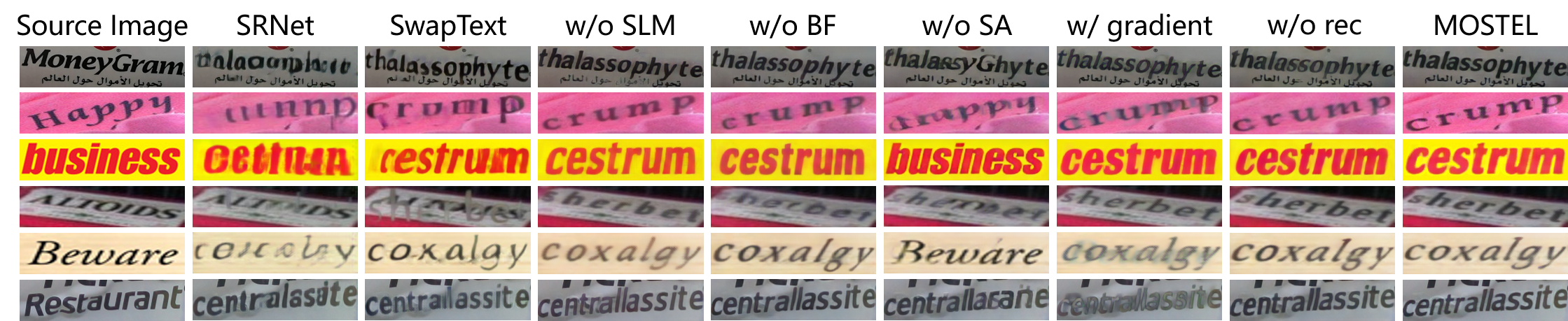}
	\caption{Some qualitative examples of previous methods and different configurations of the MOSTEL. w/o SLM, BF, SA, rec indicate MOSTEL without Stroke-Level Modification, Background Filtering, Style Augmentation, recognizer. w/ gradient means allowing the gradient propagation of the connections between decoders in BRM and TMM.}
	\label{fig:ablation}
\end{figure*}

\begin{table}[tp]
	\centering
	\renewcommand{\arraystretch}{1.2}
	\resizebox{\columnwidth}{!}
	{
		\begin{tabular}{c|cccc|c}
			\hline
			\multirow{2}*{Method} & \multicolumn{4}{c|}{Tamper-Syn2k} & Tamper-Scene \\
			\cline{2-6}
			& MSE$\downarrow$ & PSNR$\uparrow$ & SSIM$\uparrow$ & FID$\downarrow$ & SeqAcc$\uparrow$ \\
			\hline
			pix2pix\dag & 0.0732 & 12.01 & 0.3492 & 164.24 & 18.382 \\
			SRNet\dag & 0.0193 & 18.66 & 0.6098 & 41.26 & 32.298 \\
			SwapText\dag & 0.0174 & 19.43 & 0.6524 & 35.62 & 60.634 \\
			\hline
			w/o SLM & 0.0135 & 20.30 & 0.6917 & 33.59 & 70.395 \\
			w/o rec & 0.0126 & 20.50 & 0.7072 & 36.61 & 68.375 \\
			w/o BF  & 0.0134 & 20.46 & 0.7061 & 34.68 & 72.362 \\
			w/o SA & 0.0125 & 20.71 & 0.7157 & 36.50 & 26.408 \\
			w/ gradient & 0.0131 & 20.59 & 0.7105 & 38.37 & 25.670 \\
			\hline
			MOSTEL & \textbf{0.0123} & \textbf{20.81} & \textbf{0.7209} & \textbf{29.48} & \textbf{76.790} \\
			\hline
		\end{tabular}
	}
	\caption{Quantitative results on Tamper-Syn2k and Tamper-Scene. \dag means the methods that we reproduce. w/o SLM, rec, BF, SA denote without Stroke-Level Modification, recognizer, Background Filtering, Style Augmentation. w/ gradient means allowing the gradient propagation of the connections between decoders in BRM and TMM.}
	\label{tab:tab1}
\end{table}

\begin{table}[tp]
	\centering
	\renewcommand{\arraystretch}{1.2}
	\resizebox{\columnwidth}{!}
	{
		\begin{tabular}{cc|cccc|c}
			\hline
			\multirow{2}*{BRM} & \multirow{2}*{TMM} & \multicolumn{4}{c|}{Tamper-Syn2k} & Tamper-Scene \\
			\cline{3-7}
			& & MSE$\downarrow$ & PSNR$\uparrow$ & SSIM$\uparrow$ & FID$\downarrow$ & SeqAcc$\uparrow$ \\
			\hline
			- & - & 0.0124 & 20.77 & 0.7185 & 29.86 & 74.990 \\
			$\checkmark$ & - & \textbf{0.0123} & \textbf{20.81} & \textbf{0.7209} & \textbf{29.48} & \textbf{76.790} \\
			- & $\checkmark$ & 0.0130 & 20.66 & 0.7119 & 32.41 & 66.848 \\
			$\checkmark$ & $\checkmark$ & 0.0126 & 20.72 & 0.7160 & 30.66 & 70.032 \\
			\hline
		\end{tabular}
	}
	\caption{Ablation experiments of whether adding the results of SLM into training process.}
	\label{tab:slm}
\end{table}

In this section, we conduct an ablation study on Tamper-Syn2k and Tamper-Scene to show the effectiveness of our proposed Pre-Transformation, Stroke-Level Modifications and Semi-supervised Hybrid Learning.
We also provide some qualitative examples in Fig. \ref{fig:ablation}.

\subsubsection{Stroke-Level Modifications:}
After removing SLM, there is no explicit editing guidance.
Implicitly guiding the model to learn the reconstruction rules of both background and text regions undoubtedly increases the learning difficulty.
Especially at the edge of text regions, the network is ambiguous in preserving the textures or applying the editing rules, resulting in poor representation of the generated text.
The visualizations in Fig. \ref{fig:slm} and Fig. \ref{fig:ablation} show that previous methods without SLM fail to generate clear and legible characters.
The examples in Fig. \ref{fig:slm} indicate precise instructions of guidance maps.
The results in Tab. \ref{tab:tab1} also quantitatively verify the effectiveness of the proposed SLM.

Furthermore, we conduct an experiment on whether to apply SLM in the training process.
As shown in Tab. \ref{tab:slm}, when we only adopt the results of SLM in BRM to training, the model has the best performance.
We summarize the reasons as follows.
First, SLM aims to explicitly guide the network to focus on editing rules of text regions and avoid the distractions from invariant background regions.
In TMM, background regions are filtered out in advance in Pre-Transformation, so SLM is not particularly crucial here.
Second, adopting SLM in TMM discards the background regions of the feature-level fused images $\hat{O}_t$ and directly inherits from $O_s$.
Supervision on $O_t$ is actually only the supervision on text regions.
Such an image-level combination ignores the integrity between background and text, resulting in poor smoothness at the edge of text.
According to Fig. \ref{fig:slm}, compared with the image-level fusion, the feature-level fusion can integrate the background and targeted text in a more harmonious way, resulting in a smooth and seamless presentation of characters.
Therefore, we only adopt SLM in TMM at the inference stage.

\begin{table}[tp]
	\centering
	\renewcommand{\arraystretch}{1.1}
	\resizebox{\columnwidth}{!}
	{
		\begin{tabular}{c|cccc|c}
			\hline
			{Real Data} & \multicolumn{4}{c|}{Tamper-Syn2k} & Tamper-Scene \\
			\cline{2-6}
			Number & MSE$\downarrow$ & PSNR$\uparrow$ & SSIM$\uparrow$ & FID$\downarrow$ & SeqAcc$\uparrow$ \\
			\hline
			0 & \textbf{0.0122} & \textbf{20.87} & \textbf{0.7221} & \textbf{28.73} & 68.906 \\
			1 & 0.0123 & 20.83 & 0.7210 & 28.94 & 75.353 \\
			2 & 0.0123 & 20.81 & 0.7209 & 29.48 & \textbf{76.790} \\
			4 & 0.0125 & 20.76 & 0.7182 & 30.20 & 68.220 \\
			\hline
			2 (SRNet\dag) & 0.2045 & 9.71 & 0.3642 & 151.41 & 7.361 \\
			2 (Swaptext\dag) & 0.2010 & 10.61 & 0.3976 & 108.98 & 9.682\\
			\hline
		\end{tabular}
	}
	\caption{Performance about the number of real-world scene text images when using the proposed Semi-supervised Hybrid Learning. The total batch size is set to 16. \dag means that we reproduce the methods using the same configuration.}
	\label{tab:tab2}
\end{table}

\subsubsection{Semi-supervised Hybrid Learning.}
Several experiments about the number of real data in a batch have been conducted in Tab. \ref{tab:tab2}, where the total batch size is set to 16.
When the network is trained only with synthetic data, all loss functions in Fig. \ref{fig:scheme} work properly and each module can get the best optimization.
The model performs best on the Tamper-Syn2k.
When trained with real data, the network sacrifices part of its supervision, such as the reconstructed background and editing guidance maps, in exchange for adaptation to real-world scene text image distributions.
The ratio of real data to synthetic data is a parameter to trade off the network's ability to edit scene text images and the ability to achieve the expected function of each module.
With the increase of real data, due to the domain gap between synthetic and real data, the performance on Tamper-Syn2k is getting worse.
Results in Tab. \ref{tab:tab2} show that when the number is set to 2, MOSTEL has the best performance on balancing these two capabilities.
Therefore, we use batch size 16 with 14 synthetic data and 2 real data in all the following experiments.

As we stated in Methodology, such a scheme by converting the text in source image to itself faces the problem that the network may degenerate into an identity mapping one.
Three countermeasures (Background Filtering, Style Augmentation, blocking gradient) are adopted to prevent it.
As Tab. \ref{tab:tab1} shows, these measures are all effective for improving the recognition results on Tamper-Scene.
However, it is noteworthy that without Style Augmentation and blocking gradient, the model performs well on Tamper-Syn2k, but causes disastrous results on Tamper-Scene.
This is because the metrics on Tamper-Syn2k (MSE, PSNR, SSIM) are only used to statistically judge the similarity between two images without considering the visual quality.
In other words, when the model severely degenerates into an identity mapping network, the text of an input image is almost unchanged except for slight modifications in some pixel values.
Although the text content in the predicted image and the ground-truth are different, they may have similar colors, positions and pixel numbers, resulting in a high statistical metric.
While the recognition results focus on reflecting the legibility of the generated text images.
Therefore, we believe that these evaluation metrics working together can better match the visual quality of edited images.

Furthermore, we reproduce previous methods~\cite{srnet,swaptext} and train them using the same semi-supervised configuration as MOSTEL.
To be specific, for real data, we remove supervisions on all the unavailable intermediate results and use the input image to supervise the final output.
The batch size is also set to 16 with 14 synthetic data and 2 real data.
The last two rows in Tab. \ref{tab:tab2} show that they are incapable of being semi-supervised trained with both synthetic and real data, which further verifies the robustness of our structure. 

\subsubsection{Recognizer.}
As indicated in Tab. \ref{tab:tab1}, directly adopting a recognizer in training is proven to be beneficial to all the metrics.
The visualization examples in Fig. \ref{fig:ablation} also show that the model is able to generate clearer and more legible with the supervision of recognition results.

\subsection{Comparisons with Previous Methods}
To our knowledge, the code and evaluation datasets of SRNet~\cite{srnet} and Swaptext~\cite{swaptext} are not publicly available so far.
Therefore, we reproduce these methods and train them with the same training datasets and iterations as MOSTEL.
Since they do not support training with real data, which is proved in Ablation Study, we only use synthetic data to train them. 
As shown in Tab. \ref{tab:tab1}, our MOSTEL outperforms them by at least 0.0051, 1.38, 0.0685 and 6.14 in MSE, PSNR, SSIM and FID respectively on Tamper-Syn2k.
On Tamper-Scene, the recognition accuracy of MOSTEL surpasses them by over 16.156\%, demonstrating the superiority of our proposed method.

\section{Conclusion}
This study proposes an end-to-end trainable framework named MOSTEL for scene text editing.
We attribute the limited performance to implicit editing guidance and the domain gap between synthetic training data and real scene text images.
Therefore, by introducing the Stroke-Level Modification, we propose to filter out the distractions from invariant background regions and explicitly guide the model to focus on the editing rules of text regions. 
In addition, a Semi-supervised Hybrid Learning is proposed to enable the network to be trained with both paired synthetic images and unlabeled real-world images.
Several measures, such as Background Filtering, Style Augmentation and gradient-free connections, are introduced to adapt the model to such a scheme.
Extensive quantitative experiments and qualitative results verify the superiority of our method.
Besides MOSTEL, we also release two evaluation datasets named Tamper-Syn2k and Tamper-Scene to facilitate a fair comparison.
As for future work, we will combine our MOSTEL with a scene text detection network for an end-to-end text editing system and further improve the performance.

\section{Acknowledgments}
This work was supported in part by the National Key Research and Development Program of China under Grant 2022YFB3104700, and in part by the National Nature Science Foundation of China under Grants 62121002, U1936210, 61972105, and 62102384.

\bibliography{aaai23}

\end{document}